\documentclass[11pt]{article}
\usepackage{epsfig}
\usepackage{acl2001,times}

\title{Applying Natural Language Generation to Indicative Summarization
}

\author{Min-Yen Kan \and Kathleen R. McKeown \\
  Department of Computer Science \\
  Columbia University \\
  New York, NY  10027, USA \\
  \tt{\{min,kathy\}@cs.columbia.edu} \And
Judith L. Klavans \\
  Columbia University \\
  Center for Research on Information Access \\
  New York, NY, 10027 \\
  \tt{klavans@cs.columbia.edu}
}

\date{}

\begin{document}
\maketitle
\begin{abstract}
The task of creating indicative summaries that help a searcher decide
whether to read a particular document is a difficult task.  This paper
examines the indicative summarization task from a generation
perspective, by first analyzing its required content via published
guidelines and corpus analysis.  We show how these summaries can be
factored into a set of document features, and how an implemented
content planner uses the topicality document feature to create
indicative multidocument query-based summaries.
\end{abstract}

%
%

\section{Introduction}


Automatic summarization techniques have mostly neglected the {\it
indicative} summary, which characterizes what the documents are about.
This is in contrast to the informative summary, which serves as a
surrogate for the document.  Indicative multidocument summaries are an
important way of helping a user discriminate between several documents
returned by a search engine.


Traditional summarization systems are primarily based on text
extraction techniques.  For an indicative summary, which typically
describes the topics and structural features of the summarized
documents, these approaches can produce summaries that are too
specific. In this paper, we propose a natural language generation
(NLG) model for the automatic creation of indicative multidocument
summaries.  Our model is based on the values of high-level document
features, such as its distribution of topics and media types.

Specifically, we focus on the problem of content planning in
indicative multidocument summary generation.  We address the problem
of ``what to say'' in Section \ref{s:analysisBegin}, by examining what
document features are important for indicative summaries, starting
from a single document context and generalizing to a multidocument,
query-based context.  This yields two rules-of-thumb for guiding
content calculation: 1) reporting differences from the norm and 2)
reporting information relevent to the query.  

\begin{figure}
\centering
\epsfig{file=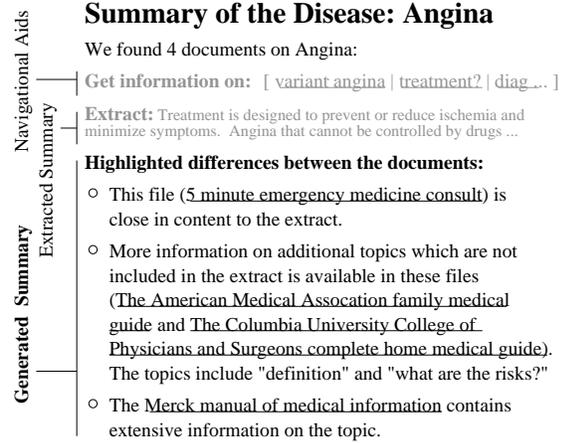,width=3.0in}
\caption{A {\sc Centrifuser} summary on the healthcare topic of
``Angina''.  The generated indicative summary in the bottom half
categorizes documents by their difference in topic distribution.}
\label{f:realSummary}
\end{figure}

We have implemented these rules as part of the content planning module
of our {\sc Centrifuser} summarization system.  The summarizer's
architecture follows the consensus NLG architecture \cite{Reiter94},
including the stages of content calculation and content planning.  We
follow the generation of a sample indicative multidocument query-based
summary, shown in the bottom half of Figure \ref{f:realSummary},
focusing on these two stages in the remainder of the paper.

\section{Document features as potential summary content}
\label{s:analysisBegin}

Information about topics and structure of the document may be based on
higher-level document features. Such information typically does not
occur as strings in the document text.  Our approach, therefore, is to
identify and extract the document features that are relevant for
indicative summaries. These features form the potential content for
the generated summary and can be represented at a semantic level in
much the same way as input to a typical language generator is
represented.  In this section, we discuss the analysis we did to
identify features of individual and sets of multiple documents that
are relevant to indicative summaries and show how feature selection is
influenced by the user query.

\subsection{Features of individual documents}

Document features can be divided into two simple categories: a) those
which can be calculated from the document body (e.g. topical structure
\cite{Hearst93} or readability using Flesch-Kincaid or SMOG
\cite{McLaughlin69} scores), and b) ``metadata'' features that may not
be contained in the source article at all (e.g. author name, media
format, or intended audience).  To decide which of these document
features are important for indicative summarization, we examined the
problem from two points of view.  From a top-down perspective, we
examined prescriptive guidelines for summarization and indexing.  We
analyzed a corpus of indicative summaries for the alternative
bottom-up perspective.

{\bf Prescriptive Guidelines}.  Book catalogues index a number of
different document features in order to provide enhanced search
access.  The United States MARC format \shortcite{MARC00}, provides
index codes for document-derived features, such as for a document's table of
contents.  It provides a larger amount of index codes for metadata
document features such as fields for unusual format, size, and special
media.  ANSI's standard on descriptions for book jackets
\shortcite{ANSI79} asks that publishers mention unusual formats,
binding styles, or whether a book targets a specific audience.

{\bf Descriptive Analysis}.  Naturally indicative summaries can also
be found in library catalogs, since the goal is to help the user find
what they need.  We extracted a corpus of single document summaries of
publications in the domain of consumer healthcare, from a local
library.  The corpus contained 82 summaries, averaging a short 2.4
sentences per summary.  We manually identified several document
features used in the summaries and characterized their percentage
appearance in the corpus, presented in Table \ref{t:documentFeatures}.

\begin{table}[hbt]
\centering
\begin{tabular}{|l|r|}
\hline
Document Feature 	& \% appearance \\
			& in corpus \\
\hline
\hline
\multicolumn{2}{|l|}{\bf Document-derived features} \\
\hline
\vspace{-.2cm}
Topicality	& 100\% \\
{\scriptsize (e.g. ``Topics include symptoms, ...'')} & \\ 
\vspace{-.2cm}
Content Types 	& 37\% \\
{\scriptsize (e.g. ``figures and tables'')} & \\ 
\vspace{-.2cm}
Internal Structure & 17\% \\
{\scriptsize (e.g. ``is organized into three parts'')} & \\ 
\vspace{-.2cm}
Readability	& 18\% \\
{\scriptsize (e.g. ``in plain English'')} & \\ 
\vspace{-.2cm}
Special Content	& 7\% \\
{\scriptsize (e.g. ``Offers 12 credit hours'')} & \\ 
Conclusions	& 3\% \\
\hline
\hline
\multicolumn{2}{|l|}{\bf Metadata features} \\
\hline 
Title		& 32\% \\
Revised/Edition & 28\% \\
Author/Editor	& 21\% \\
Purpose		& 18\% \\
Audience 	& 17\% \\
Background/Lead & 11\% \\
\vspace{-.2cm}
Source		& 8\% \\
{\scriptsize (e.g. ``based on a report'')} & \\ 
\vspace{-.2cm}
Media Type	& 5\% \\
{\scriptsize (e.g. ``Spans 2 CDROMs'')} & \\ 
\hline
\end{tabular}
\caption{Distribution of document features in library catalog
summaries of consumer healthcare publications.}
\label{t:documentFeatures}
\end{table}

Our study reports results for a specific domain, but we feel that some
general conclusions can be drawn.  Document-derived features are most
important (i.e., most frequently occuring) in these single document
summaries, with direct assessment of the topics being the most
salient.  Metadata features such as the intended audience, and the
publication information (e.g. edition) information are also often
provided (91\% of summaries have at least one metadata feature when
they are independently distributed).

\subsection{Generalizing to multiple documents}

We could not find a corpus of indicative multidocument summaries to
analyze, so we only examine prescriptive guidelines for multidocument
summarization.

The Open Directory Project's (an open source Yahoo!-like directory)
editor's guidelines \shortcite{ODP00} states that category pages that
list many different websites should ``make clear what makes a site
different from the rest''.  ``the rest'' here can mean several things,
such as ``rest of the documents in the set to be summarized'' or ``the
rest of the documents in the collection''.  We render this as the
following rule-of-thumb \ref{i:rule1}:

\begin{enumerate}
\item for a multidocument summary, a content planner should report
differences in the document that deviate from the norm for the
document's type.
\label{i:rule1}
\end{enumerate}

This suggests that the content planner has an idea of what values of a
document feature are considered normal.  Values that are significantly
different from the norm could be evidence for a user to select or
avoid the document; hence, they should be reported.  For example,
consider the document-derived feature, {\it length}: if a document in
the set to be summarized is of significantly short length, this fact
should be brought to the user's attention.

We determine a document feature's norm value(s) based on all similar
documents in the corpus collection.  For example, if all the documents
in the summary set are shorter than normal, this is also a fact that
may be significant to report to the user.  The norms need to be
calculated from only documents of similar type (i.e. documents of the
same domain and genre) so that we can model different value thresholds
for different kinds of documents.  In this way, we can discriminate
between ``long'' for consumer healthcare articles (over 10 pages)
versus ``long'' for mystery novels (over 800 pages).

\subsection{Generalizing to interactive queries}
\label{s:analysisEnd}

If we want to augment a search engine's ranked list with an indicative
multidocument summary, we must also handle queries.  The search engine
ranked list does this often by highlighting query terms and/or by
providing the context around a query term.  Generalizing this behavior
to handling multiple documents, we arrive at rule-of-thumb
\ref{i:rule2}.

\begin{enumerate}
\setcounter{enumi}{1}
\item for a query-based summary, a content planner should highlight
differences that are relevant to the query.
\label{i:rule2}
\end{enumerate}

This suggests that the query can be used to prioritize which
differences are salient enough to report to the user.  The query may
be relevant only to a portion of a document; differences outside of
that portion are not relevant.  This mostly affects document-derived
document features, such as topicality.  For example, in the consumer
healthcare domain, a summary in response to a query on treatments of a
particular disease may not want to highlight differences in the
documents if they occur in the symptoms section.

\section{Introduction to {\sc Centrifuser}}
\label{s:implementBegin}

\begin{figure}[hbt]
\centering
\epsfig{file=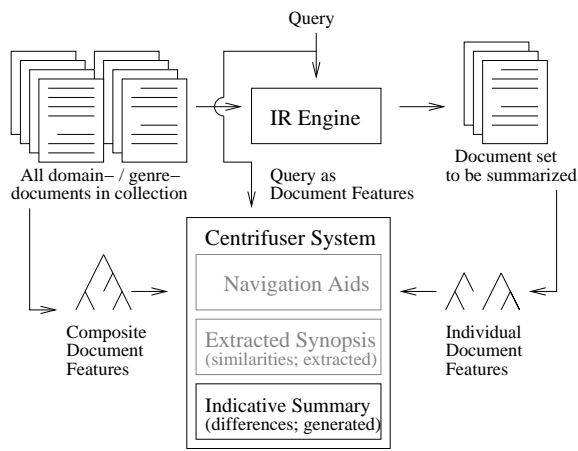,width=3.0in}
\caption{{\sc Centrifuser} architecture.}
\label{f:centrifuserArchitecture}
\end{figure}

{\sc Centrifuser} is the indicative multi-document summarization
system that we have developed to operate on domain- and genre-specific
documents.  We are currently studying consumer healthcare articles
using it.  The system produces a summary of multiple documents based
on a query, producing both an extract of similar sentences (see
\newcite{Hatzivassiloglouetal01}) as well as generating text to
represent differences.  We focus here only on the content planning
engine for the indicative, difference reporting portion.  Figure
\ref{f:centrifuserArchitecture} shows the architecture of the system.

We designed {\sc Centrifuser}'s input based on the requirements from
our analysis; document features are extracted from the input texts and
serve as the potential content for the generated summary. {\sc
Centrifuser} uses a plan to select summary content, which was
developed based on our analysis and the resulting previous rules.

Our current work focuses on the document feature which most influences
summary content and form, {\it topicality}.  It is also the most
significant and useful document feature.  We have found that
discussion of topics is the most important part of the indicative
summary.  Thus, the text plan is built around the topicality document
feature and other features are embedded as needed.  Our discussion now
focuses on how the topicality document feature is used in the system.

In the next sections we detail the three stages that {\sc Centrifuser}
follows to generate the summary: content calculation, planning and
realization.  In the first, potential summary content is computed by
determining input topics present in the document set.  For each topic,
the system assesses its relevance to the query and its prototypicality
given knowledge about the topics covered in the domain.  More
specifically, each document is converted to a tree of topics and each
of the topics is assigned a topic type according to its relationship
to the query and to its normative value.  In the second stage, our
content planner uses a text plan to select information for inclusion
in the summary.  In this stage, {\sc Centrifuser} determines which of
seven document types each document belongs to, based on the
relevance of its topics to the query and their prototypicality.  The
plan generates a separate description for the documents in each
document type, as in the sample summary in Figure \ref{f:realSummary},
where three document categories was instantiated.  In the final stage,
the resulting description is lexicalized to produce the summary.

\section{Computing potential content: topicality as topic trees}
\label{s:topicTrees}

\begin{figure}
\centering
\epsfig{file=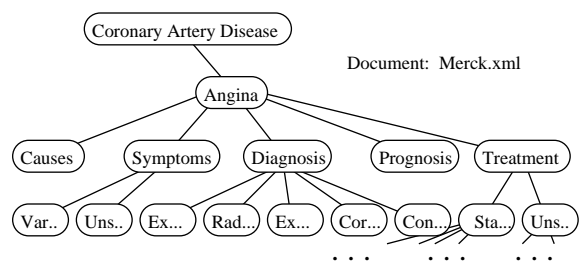,width=3.0in}
\caption{A topic tree for an article about coronary artery disease
from {\it The Merck manual of medical information}, constructed
automatically from its section headers.}
\label{f:topicTree}
\end{figure}

In {\sc Centrifuser}, the topicality document feature for individual
documents is represented by a tree data structure.  Figure
\ref{f:topicTree} gives an example {\it document topic tree} for a
single consumer healthcare article.  Each document in the collection
is represented by such a tree, which breaks each document's topic into
subtopics.

We build these document topic trees automatically for structured
documents using a simple approach that utilizes section headers, which
suffices for our current domain and genre.  Other methods such as
layout identification \cite{HuEtAl99} and text segmentation /
rhetorical parsing \cite{Yaari99,KanEtal98,Marcu97} can serve as the
basis for constructing such trees in both structured and unstructured
documents, respectively.

\subsection{Normative topicality as composite topic trees}

As stated in rule \ref{i:rule1}, the summarizer needs normative values
calculated for each document feature to properly compute differences
between documents.  

\begin{figure}
\centering
\epsfig{file=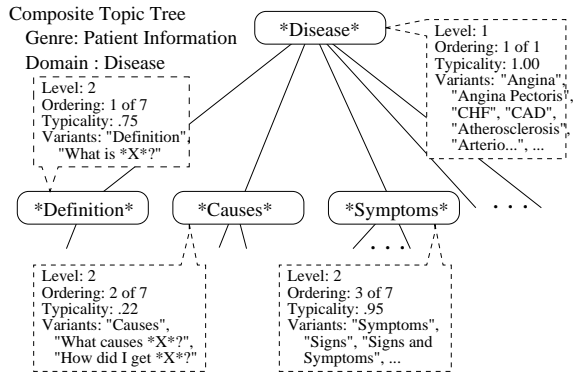,width=3.0in}
\caption{A sample composite topic tree for consumer health information
for diseases.}
\label{f:compositeTopicTree}
\end{figure}

The {\it composite topic tree} embodies this paradigm.  It is a data
structure that compiles knowledge about all possible topics and their
structure in articles of the same intersection of domain and genre,
(i.e., rule \ref{i:rule1}'s notion of ``document type'').  Figure
\ref{f:compositeTopicTree} shows a partial view of such a tree
constructed for consumer healthcare articles.

The composite topic tree carries topic information for all articles of
a particular domain and genre combination. It encodes each topic's
relative typicality, its prototypical position within an article, as
well as variant lexical forms that it may be expressed as
(e.g. alternate headers).  For instance, in the composite topic tree
in Figure \ref{f:compositeTopicTree}, the topic ``Symptoms'' is very
typical (.95 out of 1), may be expressed as the variant ``Signs'' and
usually comes after other its sibling topics (``Definition'' and
``Cause'').

Compiling composite topic trees from sample documents is a non-trivial
task which can be done automatically given document topic trees.
Within our project, we developed techniques that align multiple
document topic trees using similarity metrics, and then merge the
similar topics \cite{KanEtal01-CTT}, resulting in a composite
topic tree.

\section{Content Planning} 
\label{s:contentPlanning}

NLG systems traditionally have three components: content planning,
sentence planning and linguistic realization.  We will examine how the
system generates the summary shown earlier in Figure
\ref{f:realSummary} by stepping through each of these three steps.

\begin{figure}[hbt]
\centering
\epsfig{file=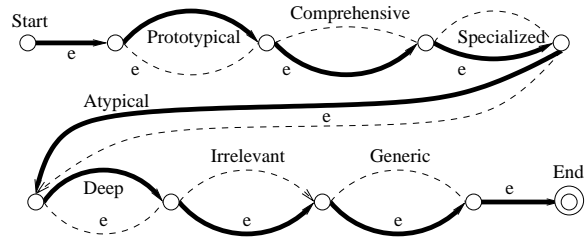,width=3in}
\label{f:intercategoryDiscoursePlan}
\caption{Indicative summary content plan, solid edges indicate moves
in the sample summary.}
\end{figure}

During content planning, the system decides what information to convey
based on the calculated information from the previous stage.  Within
the context of indicative multidocument summarization, it is important
to show the differences between the documents (rule \ref{i:rule1}) and
their relationship to the query (rule \ref{i:rule2}).  One way to do
so is to classify documents according to their topics' prototypicality
and relevance to the query.  Figure \ref{f:intercategoryDiscoursePlan}
gives the different document categories we use to capture these
notions and the order in which information about a category should be
presented in a summary.

\subsection{Document categories}

Each of the document categories in the content plan in Figure
\ref{f:intercategoryDiscoursePlan} describes documents that are
similar in their distribution of information with respect to the
topical norm (rule \ref{i:rule1}) and to the query (rule
\ref{i:rule2}).  We explain these {\it document categories} found in
the text plan below.  The examples in the list below pertain to a
general query of ``Angina'' (a heart disorder) in the same domain of
consumer healthcare.

{\bf 1. Prototypical} - contains information that one would typically
expect to find in an on-topic document of the domain and genre.  An
example would be a reference work, such as {\it The AMA Guide to
Angina}.

{\bf 2. Comprehensive} - covers most of the typical content but may
also contain other added topics.  An example could be a chapter of a
medical text on angina.

{\bf 3. Specialized} - are more narrow in scope than the previous two
categories, treating only a few normal topics relevant to the query.
A specialized example might be a drug therapy guide for angina.

{\bf 4. Atypical} - contains high amounts of rare topics, such as
documents that relate to other genres or domains, or which discuss
special topics.  If the topic ``Prognosis'' is rare, then a document
about life expectancy of angina patients would be an example.

{\bf 5. Deep} - are often barely connected with the query topic but
have much underlying information about a particular subtopic of the
query.  An example is a document on ``Surgical treatments of Angina''.

{\bf 6. Irrelevant} - contains mostly information not relevant to the
query. The document may be very broad, covering mostly unrelated
materials.  A document about all cardiovascular diseases may be
considered irrelevant.

{\bf 7. Generic} - don't display tendencies towards any particular
distribution of information.

\subsection{Topic types}

Each of these document categories is different because they have an
underlying difference in their distribution of information.  {\sc
Centrifuser} achieves this classification by examining the
distribution of {\it topic types} within a document.  {\sc
Centrifuser} types each individual topic in the individual document
topic trees as one of four possibilities: {\it typical}, {\it rare},
{\it irrelevant} and {\it intricate}.  Assigning topic types to each
topic is done by operationalizing our two content planning rules.

\begin{figure}
\centering
\epsfig{file=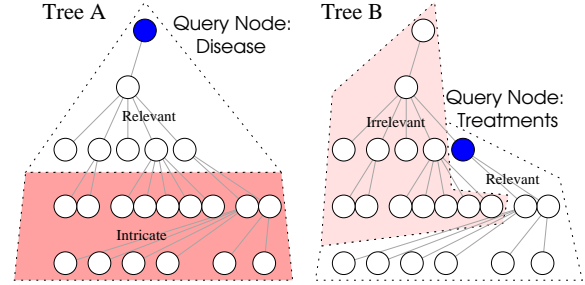,width=3.0in}
\caption{The three topic regions as defined by the query, for $k$ = 2
($k$ being the {\it intricate} beam depth.}
\label{f:topicTypes}
\end{figure}

To apply rule \ref{i:rule2}, we map the text query to the single most
similar topic in each document topic tree (currently done by string
similarity between the query text and the topic's possible lexical
forms).  This single topic node -- the query node -- establishes a
relevant scope of topics.  The relevant scope defines three regions in
the individual topic tree, shown in Figure \ref{f:topicTypes}: topics
that are relevant to the query, ones that are too {\it intricate}, and
ones that are {\it irrelevant} with respect to the query.  Irrelevant
topics are not subordinate to the query node, representing topics that
are too broad or beyond the scope of the query.  Intricate topics are
too detailed; they are topics beyond $k$ hops down from the query
node.

Each individual document's ratio of topics in these three regions thus
defines its relationship to the query:  a document with mostly
information on treatment would have a high ratio of relevant to other
topics if given a treatment query; but the same document given a query
on symptoms would have a much lower ratio.

To apply rule \ref{i:rule1}, we need to know whether a particular
topic ``deviates from the norm'' or not.  We interpret this as whether
or not the topic normally occurs in similar documents -- exactly the
information encoded in the composite topic tree's typicality score.
As each topic in the document topic trees is an instance of a node in
the composite topic tree, each topic can inherit its composite node's
typicality score.  We assign nodes in the relevant region (as defined
by rule 2), with labels based on their typicality.  For convenience,
we set a typicality threshold $\alpha$, above which a topic is
considered {\it typical} and below which we consider it {\it rare}.

At this point each topic in a document is labeled as one of the four
topic types.  The distribution of these four types determines each
document's document category.  Table \ref{t:distribution} gives the
distribution parameters which allow {\sc Centrifuser} to classify the
documents.

\begin{table}[hbt]
\centering
\small
\begin{tabular}{|l|l|}
\hline
Document Category 	& Topic Distribution \\
\hline

\hline
1. Prototypical 		& $>$ 50+\% {\it typical} and \\
				& $>$ 50+\% all possible {\it typical} \\
2. Comprehensive		& $>$ 50+\% all possible {\it typical} \\
3. Specialized			& $>$ 50+\% {\it typical} \\
4. Atypical			& $>$ 50+\% {\it rare} \\
5. Deep				& $>$ 50+\% {\it intricate} \\
6. Irrelevant			& $>$ 50+\% {\it irrelevant} \\
7. Generic			& n/a \\
\hline
\end{tabular}
\caption{Classification rules for document categories.}
\label{t:distribution}
\end{table}

Document categories add a layer of abstraction over the topic types
that allow us to reason about documents.  These document labels still
obey our content planning rules \ref{i:rule1} and \ref{i:rule2}: since
the assignment of a document category to a document is conditional on
its distribution of its topics among the topic types, a document's
category may shift if the query or its norm is changed.

In {\sc Centrifuser}, the text planning phase is implicitly performed
by the classification of the summary document set into the document
categories.  If a document category has at least one document
attributed to it, it has content to be conveyed.  If the document
category does not have any documents attributed to it, there is no
information to convey to the user concerning the particular category.

An instantiated document category conveys a couple of messages.  A
description of the document type as well as the elements attributed to
it constitutes the minimal amount of information to convey.  Optional
information such as details on the instances, sample topics or other
unusual document features, can be expressed as well.

\begin{figure}[hbt]
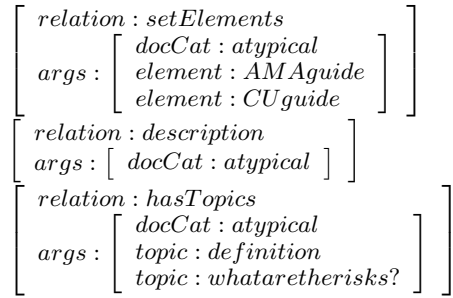

\small
$\left [
\begin{array}{l}
relation: setElements \\
args: \left [ 
  \begin{array}{l}
    docCat: atypical \\
    element: AMA guide \\
    element: CU guide \\
  \end{array}
  \right ] \\
\end{array}
\right ] $
$\left [
\begin{array}{l}
relation: description \\
args: \left [ 
  \begin{array}{l}
    docCat: atypical \\
  \end{array}
  \right ] \\
\end{array}
\right ] $
$\left [
\begin{array}{l}
relation: hasTopics \\
args: \left [ 
  \begin{array}{l}
    docCat: atypical \\
    topic: definition \\
    topic: what are the risks? \\
  \end{array}
  \right ] \\
\end{array}
\right ] $
\caption{Messages instantiated for the {\it atypical} document
category for the summary in Figure \ref{f:realSummary}.}
\end{figure}

The text planner must also order the selected messages into a coherent
plan for subsequent realization.  For our summary, this is a problem
on two levels: deciding the ordering between the document category
descriptions and deciding the ordering of the individual messages
within the document category.  In {\sc Centrifuser}, the discourse
plans for both of these levels are fixed.  Let us first discuss the
inter-category plan.

{\bf Inter-category.}  We order the document category descriptions
based on the ordering expressed in Table \ref{t:distribution}.  The
reason for this order is partially reflected by the category's
relevance to the user query (rule \ref{i:rule2}).  Document categories
like {\it prototypical} whose salient feature is their high ratio of
relevant topics, are considered more important than document
categories that are defined by their ratio of intricate or irrelevant
topics (e.g. {\it deep}).

This precendence rule decides the ordering for the last few document
types (deep $\rightarrow$ irrelevant $\rightarrow$ generic).  For the
remaining document types, defined by their high ratio of typical and
rare topics, we use an additional constraint of ordering document
types that are closer to the article type norm before others.  This
orders the remaining beginning topics (prototypical $\rightarrow$
comprehensive $\rightarrow$ specialized $\rightarrow$ atypical).  The
reason for this is that {\sc Centrifuser}, along with reporting
salient differences by using NLG, also reports an multidocument
extract based on similarities.  As similarities are drawn mostly from
common topics -- that is, {\it typical} ones -- typical topics are
regarded as more important than rare ones.  

Figure \ref{f:intercategoryDiscoursePlan} shows the resulting
inter-category discourse plan.  As stated in the text planning phase,
if no documents are associated with a particular document category, it
will be skipped, reflected in the figure by the $\epsilon$ moves.  Our
sample summary summary contains prototypical (first bullet), atypical
(second) and deep (third) document categories, and as such activates
the solid edges in the figure.

\begin{figure}[hbt]
\centering
\epsfig{file=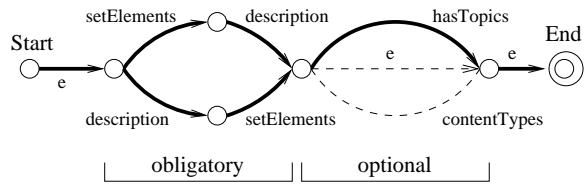,width=3in}
\caption{Intra-category discourse plan, solid edges indicate moves in
the atypical document category.  The final choice on which obligatory
structure to use is decided later during realization.}
\label{f:intracategoryDiscoursePlan}
\end{figure}

{\bf Intra-category.}  Ordering the messages within a category follows
a simple rule.  Obligatory information is expressed first, while
optional information is expressed afterwards.  Thus the document
category's constituents and its description always come first, and
information about sample topics or other unusual document features
come afterwards, shown in Figure \ref{f:intracategoryDiscoursePlan}.
The result is a partial ordering (as the order of the messages in the
obligatory information has not been fixed) that is linearized later.

\section{Sentence Planning and Lexical Choice}

In the final step, the discourse plan is realized as text.  First, the
sentence planner groups messages into sentences and generates
referring expressions for entities.  Lexical choice also happens at
this stage.  In our generation task, the grouping task is minimal; the
separate categories are semantically distinct and need to be realized
separately (e.g., in the sample, each category is a separate list
item).  The obligatory information of the description of the category
as well as the members of the category are combined into a single
sentence, and optional information (if realized) constitute another
sentence.

\subsection{Generating Referring Expressions}

One concern for generating referring expressions is constraining the
size of the sentence.  This is an issue when constructing referring
expressions to sets of documents matching a document type.  For
example, if a particular document category has more than five
documents, listing the names of each individual document is not
felicitous.  In these cases, an exemplar file is picked and used to
demonstrate the document type.  Resulting text is often of the form:
``There are 23 documents (such as the {\it AMA Guide to Angina}) that
have detailed information on a particular subtopic of angina.''

Another concern in the generation of referring expressions is when the
optional information only applies to a subset of the documents of the
category.  In these cases, the generator will reorder the elements of
the document category in such a way to make the subsequent referring
expression more compact (e.g. ``The first five documents contain
figures and tables as well'' versus the more voluminous ``The first,
third, fifth and the seventh documents contain figures and
tables as well'').

\begin{figure}[hbt]
\centering \small
\begin{verbatim}
(S1/description+setElements
  (V1 :value ``be available'')
  (NP1/atypical :value 
     ``more information on additional 
       topics which are not included 
       in the extract'')
  (NP2/setElements :value 
     ``files (The AMA guide and 
       CU Guide)''))
(S2/hasTopics
  (V1 :value ``include'')
  (NP1/atypicalTopics :value ``topics'')
  (NP2/topicList :value 
     ``definition and 
       what are the risks?''))
\end{verbatim}
\caption{Sentence plan for the atypical document category.}
\label{f:sentencePlan}
\end{figure}

\subsection{Lexical Choice}

Lexical choice in {\sc Centrifuser} is performed at the phrase level;
entire phrases can be chosen all at once, akin to template based
generation.  Currently, a path is randomly chosen to select a
lexicalization.  In the sample summary, the atypical document
category's (i.e. the second bullet item) description of ``more
information on additional topics ...'' was chosen as the description
message among other phrasal alternatives.  The sentence plan for this
description is shown in Figure \ref{f:sentencePlan}.

For certain document categories, a good description can involve
information outside of the generated portion of the summary.  For
instance, Figure \ref{f:realSummary}'s prototypical document category
could be described as being ``an reference document about angina''.
But as a prototypical document shares common topics among other
documents, it is actually well represented by an extract composed of
the similarities across document sets.  Similarity extraction is done
in another module of {\sc Centrifuser} (the greyed out portion in the
figure), and as such we also can use a phrasal description that
directly references its results (e.g., in the actual description used
for the prototypical document category in Figure \ref{f:realSummary}).

\subsection{Linguistic Realization}

Linguistic realization takes the sentence plan and produces actual
text by solving the remaining morphology and syntactic problems.  {\sc
Centrifuser} currently chooses a valid syntactic pattern at random, in
the same manner as lexical choice.  Morphological and other agreement
constraints are minor enough in our framework and are handled by set
rules.

\section{Current status and future work}

{\sc Centrifuser} is fully implemented; it produces the sample summary
in Figure \ref{f:realSummary}.  We have concentrated on implementing
the most commonly occuring document feature, topicality, and have
additionally incorporated three other document features into our
framework (document-derived {\it Content Types} and {\it Special Content}
and the {\it Title} metadata).  

Future work will include extending our document feature analysis to
model context (to model adding features only when appropriate), as
well as incorporating additional document features.  We are also
exploring the use of stochastic corpus modeling
\cite{Langkilde00,Bangalore&Rambow00} to replace our template-based
realizer with a probabilistic one that can produce felicitous sentence
patterns based on contextual analysis.

\section{Conclusion}
\label{s:conclusion}
We have presented a model for indicative multidocument summarization
based on natural language generation. In our model, summary content is
based on document features describing topic and structure instead of
extracted text. Given these features, a generation model uses a text
plan, derived from analysis of naturally occurring indicative
summaries plus guidelines for summarization, to guide the system in
describing document topics as typical, rare, intricate, or relevant to
the user query. We showed how the topicality document feature can be
derived from the set of input documents and represented as a topic
tree for each document along with a merged composite topic for all
documents in the collection against which prototypicality and query
relevance can be computed. Our ongoing work is examining how to
automatically learn the text plans along with the tactics needed to
realize each piece of the instantiated plan as a sentence.

\bibliographystyle{acl}
\bibliography{wnlg01}
\clearpage
\end{document}